\documentclass{article}

\usepackage[final]{neurips_2021}

\usepackage[utf8]{inputenc} 
\usepackage[T1]{fontenc}    
\usepackage[colorlinks=true, citecolor=blue, linkcolor=blue, urlcolor=blue]{hyperref}       
\usepackage{url}            
\usepackage{booktabs}       
\usepackage{amsfonts}       
\usepackage{nicefrac}       
\usepackage{microtype}      
\usepackage{xcolor}         
\usepackage{amsmath}
\usepackage{graphicx}
\usepackage{natbib}
\usepackage{subcaption}

\DeclareMathOperator*{\softmax}{softmax}
\newcommand{\fm}[1]{\textit{#1}}

\title{Systematic Generalization with Edge Transformers}

\author{%
  Leon Bergen \\
  University of California, San Diego \\
  \texttt{lbergen@ucsd.edu } \\
  \And
  Timothy J. O'Donnell \\
  McGill University \\
  Quebec Artificial Intelligence Institute (Mila) \\
  Canada CIFAR AI Chair 
  \And 
  Dzmitry Bahdanau \\
  Element AI, a ServiceNow company \\
  McGill University \\
  Quebec Artificial Intelligence Institute (Mila) \\
  Canada CIFAR AI Chair 
}

\begin{document}

\maketitle

\begin{abstract}
  Recent research suggests that systematic generalization in natural
  language understanding remains a challenge for state-of-the-art
  neural models such as Transformers and Graph Neural Networks. To
  tackle this challenge, we propose Edge Transformer, a new model
  that combines inspiration from Transformers and rule-based symbolic
  AI. The first key idea in Edge Transformers is to associate vector
  states with every edge, that is, with every pair of input nodes---as
  opposed to just every node, as it is done in the Transformer model. The
  second major innovation is a \textit{triangular attention} mechanism
  that updates edge representations in a way that is inspired by 
  unification from logic programming. We evaluate Edge Transformer on
  compositional generalization benchmarks in relational reasoning, 
  semantic parsing, and dependency parsing\footnote{Code for our experiments can be found at: github.com/bergen/EdgeTransformer}. In all three settings, the Edge
  Transformer outperforms Relation-aware, Universal and classical
  Transformer baselines.
  \end{abstract}

\section{Introduction}

Transformers \citep{vaswani_attention_2017} have become ubiquitous in
natural language processing and deep learning more generally
\citep[e.g.,][]{devlin_bert_2018,raffel_exploring_2020,carion_end--end_2020}.
Nevertheless, systematic (or compositional) generalization benchmarks
remain challenging for this class of models, including large instances
with extensive pre-training
\citep{keysers_measuring_2020,tsarkov_-cfq_2020,gontier_measuring_2020,
  furrer_compositional_2020}. Similarly, despite their increasing
application to a variety of reasoning and inference problems, the
ability of Graph Neural Networks' (GNN)
\citep{gori_new_2005,scarselli_graph_2009,velickovic_graph_2018} to
generalize systematically has also been recently called into question
\citep{sinha_clutrr_2019,sinha_graphlog_2020}. Addressing these
challenges to systematic generalization is critical both for robust
language understanding and for reasoning from other kinds of knowledge
bases.

In this work, inspired by symbolic AI, we seek to equip Transformers
with additional representational and computational capacity to better
capture the kinds of information processing that underlie
systematicity. Our intuition is that classical Transformers can be
interpreted as a kind of inferential system similar to a continuous
relaxation of a subset of the Prolog logic programming language. In
particular, as we discuss in the next section, Transformers can be
seen as implementing a kind of reasoning over \fm{properties} of
entities---for example, \textsc{red}($\cdot$) or
\textsc{table}($\cdot$)---where any \textit{relations} required for
such reasoning---for example \textsc{part-of}($\cdot$, $\cdot$) or
\textsc{grandmother}($\cdot$, $\cdot$)---are inferred on the fly with
the self-attention mechanism.

Building on this intuition, in this work we propose
  \fm{Edge Transformers}, a generalization of the
  Transformer model that uses a novel \fm{triangular attention}
  mechanism that is inspired by a much more general family of
  inferential rules.
  
  To achieve this, we endow the Edge Transformer with a 3D tensor
  state such that every \textit{edge}, that is, every pair of input
  nodes, contains a vector that represents relations between the two
  nodes.  The updates of each edge are computed using all adjacent
  edges in a way that is directly inspired by unification in
  logic programming.  While the use of edge features or even dynamic
  edge states can be found in the Transformer and GNN literatures
  \citep{shaw_self-attention_2018,gilmer_neural_2017,gong_exploiting_2019},
  to the best of our knowledge such triangular updates are novel to
  our approach.

  We evaluate the Edge Transformers on three recently proposed
  compositional generalization challenges. First, on the graph version
  of the CLUTRR relational reasoning challenge
  \mbox{\citep{sinha_clutrr_2019}}, our model displays stronger
  systematic generalization than Graph Attention Networks
  \citep{velickovic_graph_2018} and Relation-aware Transformers
  \mbox{\citep{shaw_self-attention_2018}}. Second, on both dependency
  parsing and semantic parsing versions \citep{cfq_dependencies} of
  the Compositional Freebase Questions benchmark
  \citep{keysers_measuring_2020} our model achieves a higher parsing
  accuracy than that of classical Transformers and BiLSTMs. Last but
  not least, Edge Transformer achieves state-of-the-start performance
  of 87.4\% on the COGS benchmark for compositional generalization in
  semantic parsing.

 \section{Intuitions}

 A Transformer operates over a set of $n$ entities---such as words in
 a sentence---which we represent as \fm{nodes} in a graph like those
 displayed on the left-hand side of
 Figure~\ref{fig:edge_transformer}. Each node is associated with a
 $d$-dimensional \fm{node-state} $X(i)$ for $i \in n$, which can be
 thought of as the output associated with applying some function to
 the node---that, a representation of node \fm{properties}. In
 transformers, computation proceeds by sequentially associating each
 node with a number $l$ of node-states, with each state updated from
 node states at the preceding layer via the attention mechanism.

 Adopting a Prolog-like notation, we could write the fundamental
 inferential process implemented by the transformer architecture as
\begin{align}
    X^{l +1}(1) \vdash_\mathcal{A} X^{l}(1), X^{l}(2), \dots, X^{l}(n).
    \label{eq:label1}
\end{align}
In logic programming, the turnstile symbol $\vdash$
means that whenever the right-hand side of the expression is true, the
left-hand side must be true. In our interpretation of Transformers,
the inference rules expressed by $\vdash_{\mathcal{A}}$ are learned
via the attention mechanism $\mathcal{A}$, as is the meaning of each
property $X^l(\cdot)$. Classical transformers can therefore be
interpreted as an architecture specialized to learning how entity
properties can be inferred from the properties of other entities.

Despite the power of this mechanism, it still has noteworthy
limitations. In particular, for many practical reasoning and NLU tasks
systems must learn dependencies between \textit{relations}, such as family
relations: $\textsc{mother}(x,y)$ and $\textsc{mother}(y,z)$ implies
$\textsc{grandmother}(x,z)$. Such a general reasoning problem might be
expressed in Prolog-like notation as follows.
\begin{align}
    X^{l +1}(1, 2) \vdash_\mathcal{A} X^{l}(1,1),  X^{l}(1,2),  \dots, X^{l}(2,1),  X^{l}(2,2), \dots.
    \label{eq:label1}
\end{align}
It is of course possible for classical transformers to capture such
reasoning patterns. But for them to do so, each transformer state
$X^{l +1}(1)$ must encode \textbf{both} the properties of the relation
itself, that is, $\textsc{mother}(\cdot,\cdot)$ versus
$\textsc{grandmother}(\cdot,\cdot)$---as well as all target nodes $x$
with which node $1$ stands in the relation, that is
$\textsc{mother}(1,x)$. In other words, to encode relations a
classical transformer must use its state representations for two
distinct purposes. This mixing of concerns places a high burden on the
learning, and we hypothesize that the resulting inductive bias hinders
the ability of this class of models to generalize systematically.

To address this problem we propose that the fundamental object
represented by the model should not be a state associated with a node
but, rather, states associated with edges, like those displayed on the
right of Figure~\ref{fig:edge_transformer}. Such edge representations
in turn are updated based on an attention mechanism which attends to
other edges. This mechanism is inspired by the process of unification
in logic programming. Critical to an inference rule such as
$\textsc{grandmother}(x,y)~\vdash_\mathcal{A}~\textsc{mother}(x,z)~\textsc{mother}(z,y)$ is the fact that the two predicates on the right
hand side of this rule share a variable $z$. Successfully inferring
$\textsc{grandmother}(x,y)$ which involve finding a binding for $z$
which satisfies both $\textsc{mother}(z,y)$ and
$\textsc{mother}(x,z)$---a process which is handled by
\textit{unification} in logic programming. We believe that this is a
fundamental aspect of relational reasoning and to capture this we make
use of a form of attention which we call \fm{triangular attention} and
which we further describe in the next section.

\section{Edge Transformers}
\label{sec:edge_transformer}

\begin{figure}
    \centering
    \includegraphics[trim=100 200 100 200,  width=\textwidth]{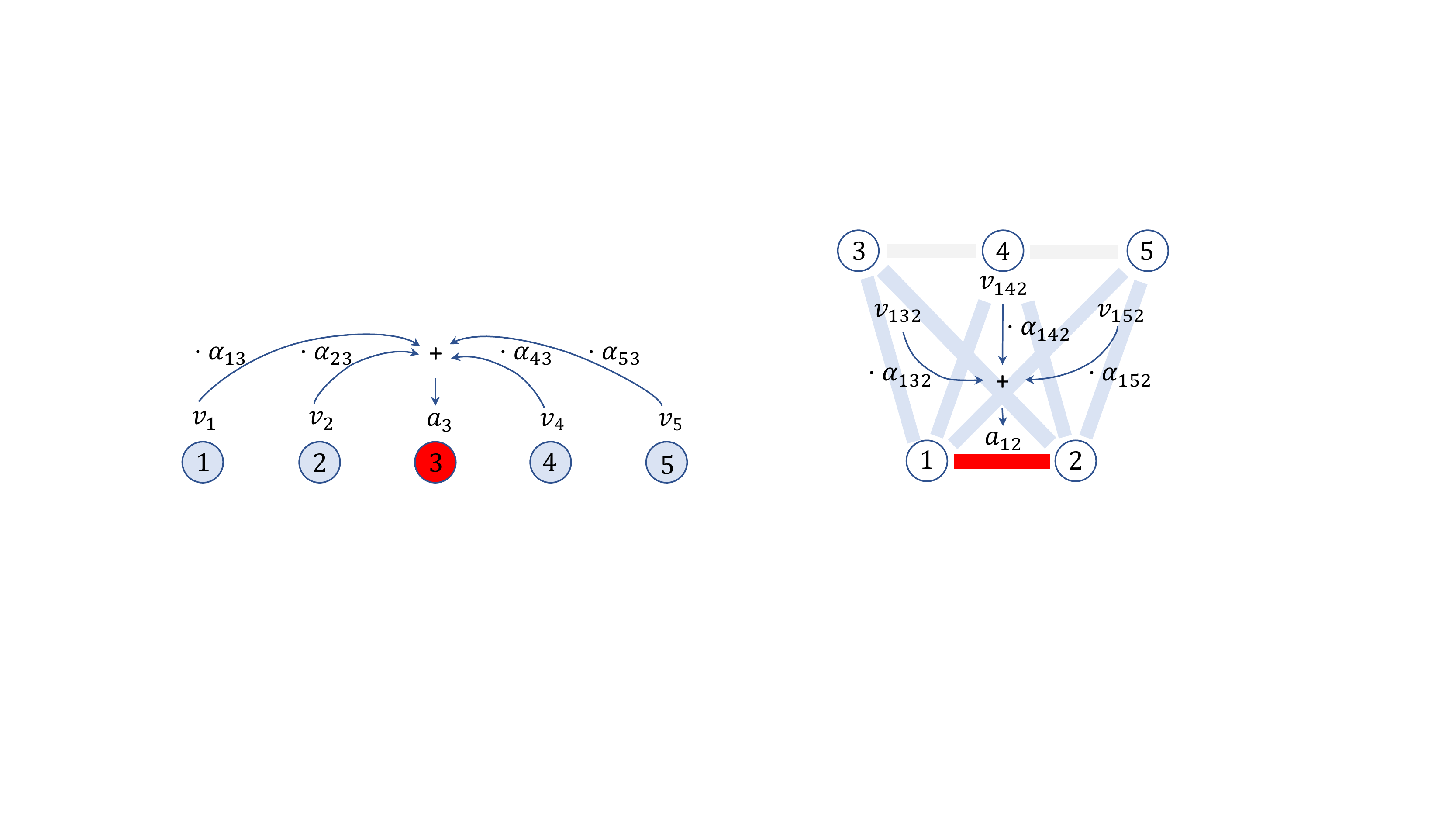}
    \caption{Left: Transformer self-attention computes an update $a_3$
      for the node $3$ by mixing value vectors $v_i$ associated with
      each node $i$ with attention weights $\alpha_i$. Right:
      triangular attention computes an update $a_{12}$ for the edge
      $(1, 2)$ by mixing value vectors $v_{1l2}$ computed for each
      triangle $(1, l, 2)$ with attention weights $\alpha_{1l2}$.
      Values $v_{3}$, $v_{112}$, $v_{122}$ and their contributions are
      not shown for simplicity.}
    \label{fig:edge_transformer}
  \end{figure}
  
An edge transformer operates on a complete graph with $n$ nodes and
$n^2$ directed edges (we allow self-connections) labeled with
$d$-dimensional feature vectors. An edge transformer state therefore
is a 3-dimensional tensor $X \in \mathbb{R}^{n, n, d}$ that consist
of edges states $X(i,j)$ corresponding to every edge $(i, j)$,
$i, j \in \left[1, n\right]$. We will also write $x_{ij}$ for each
edge state. An edge transformer layer computes
\begin{align}
  X' = \mathrm{FFN}(\textsc{ln}(X +\mathrm{TriangularAttention}(\textsc{ln}(X)))).
    \label{eq:edge_transformer}
\end{align}
Here, $\textsc{FFN}(X)$ is a fully-connected feed-forward network with
one hidden layer of $4 \cdot d$ units that is applied independently to each
edge state $x_{ij}$, $\textsc{ln}$ stands for the layer normalization
mechanism \citep{ba_layer_2016} which rescales activations of each
feature $f \in \left[1, d\right]$ across its occurrences in edge
states $x_{ij}$. Equation~\ref{eq:edge_transformer} follows the
familiar structure of transformer \citep{vaswani_attention_2017}
computations with two important differences: (a) It uses a 3D-tensor state
instead of a matrix state; and (b) it makes use of a novel triangular attention
mechanism that we describe below.

For a single edge state $x_{ij}$ a single-head triangular attention update
outputs a vector $a_{ij}$ that is computed as follows:
\begin{align}
    a_{ij} = W^o \sum\limits_{l \in \left[1, n\right]} \alpha_{ilj} v_{ilj},  \\
    \alpha_{ilj} = \softmax\limits_{l \in [1, n]} q_{il}  k_{lj} / \sqrt{d},
    \label{eq:edge_attention_begin}\\
    q_{il} = W^q x_{il}, \label{eq:query}\\
    k_{lj} = W^k x_{lj}, \label{eq:key}\\
    v_{ilj} = V^{1} x_{il} \odot V^2 x_{lj},
    \label{eq:edge_attention_end}
\end{align}
where $\odot$ stands for elementwise multiplication, and
$W^q, W^k, W^o, V^1, V^2 \in \mathbb{R}^{d,d}$ are matrices that are
used to produce key, query, output and value vectors $k_{il}$,
$q_{lj}$, $a_{ij}$ and $v_{ilj}$ respectively. Here and in the rest of
the paper we omit bias terms for clarity.

Informally, updates associated with an edge $(i,j)$ proceed by
aggregating information across all pairs of edges that share a node
$l$, that is, all pairs $(i, l)$ and $(l,j)$. The updated edge value,
$a_{ij}$, is an attention-weighted mixture of contributions from each
such pair of edges.
Figure \ref{fig:edge_transformer} visualizes some of the key
differences between transformer and edge transformer computations.
Note that edges $(i,j)$, $(i,l)$, and $(l,j)$ form a triangle in the
figure---hence the name of our attention mechanism.

We define a multi-head generalization of the above mechanism in a way
similar to the implementation of multi-head attention in vanilla
transformers.  Specifically, each head $h \in [1, m]$ will perform
Edge Attention using the smaller corresponding matrices
$W^q, W^k, V^1, V^2 \in \mathbb{R}^{d,d / m}$, and the joint output
$a_{ij}$ will be computed by multiplying a concatenation of the heads'
output by $W^o$: \begin{align} a_{ij} = W^o \left[ a^{1}_{ij}; \ldots;
    a^{m}_{ij} \right].
\end{align}
In practice a convenient way to implement an efficient batched version
of edge attention is using the Einstein summation operation which is
readily available in modern deep learning frameworks. 

\paragraph{Tied \& Untied Edge Transformer}

Edge Transformer layers can be stacked with or without tying weights
across layers. By default we tie the weights in a way similar to 
Universal Transformer \citep{dehghani_universal_2019}. We will refer
to the version with weights untied as Untied Edge Transformer.

\paragraph{Input \& Output}

The Edge Transformer's initial state $X^0$ is defined differently for
different applications. When the input is a labeled graph with
(possibly null) edge labels, the initial state $x^0_{ij}$ is produced
by embedding the labels with a trainable embedding layer. When the
input is a sequence, such as for example a sequence of words or tokens
$t_i$, self-edges $x^0_{ii}$ are initialized with the embedding for
token $t_i$, and all edges $x^0_{ij}$ are initialized with a relative
position embedding \citep{shaw_self-attention_2018}:
\begin{align} x^0_{ij} = 
\begin{cases} e(t_i) + a(0) &\mbox{if } i=j \\
a(i-j) & \mbox{if } i \neq j \end{cases}
    \label{eq:sequence_init}
\end{align}
where $e$ is a trainable
$d$-dimensional embedding layer, and $a(i - j)$ is a relative position embedding for the position difference $i - j$.

\begin{figure}
  \centering
  \includegraphics[scale=0.22]{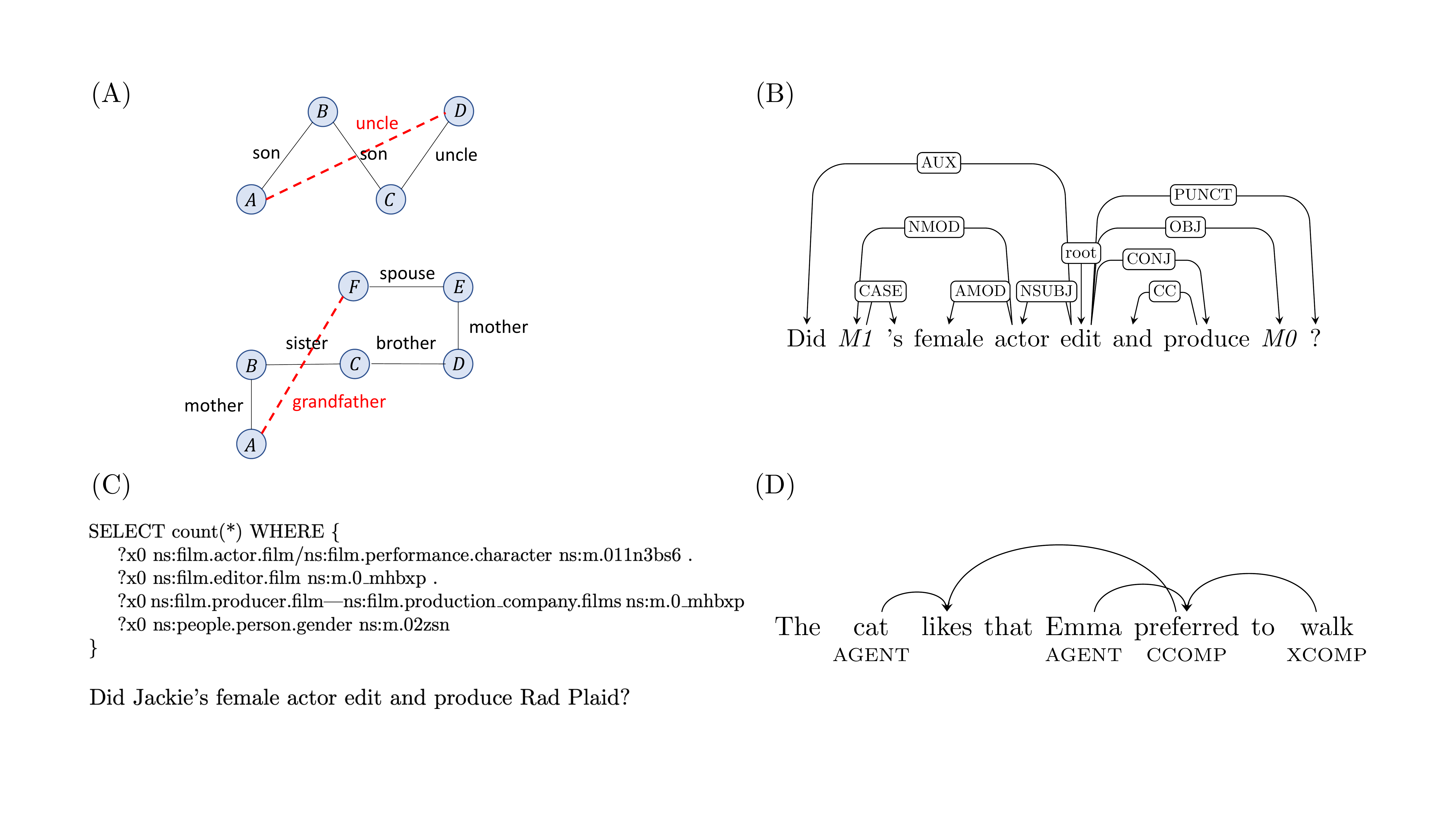}
    \caption{Illustrations of the benchmarks that we use to evaluate
      the Edge Transformer.  \textbf{(A)}~Relational reasoning
      scenarios for the CLUTRR benchmark, with the relation length of
      3 (top) and 6 (bottom).  \textbf{(B)}~Dependency parsing
      for Compositional Freebase Questions
      \citep[CFQ; reproduced with
      permission,][]{goodwin2021compositional}.  \textbf{(C)}~SPARQL query and the corresponding question from the
      CFQ semantic parsing dataset.  \textbf{(D)} graph
      representations of COGS semantic parses
      \citep{ontanon2021making}.
      \label{fig:tasks}}
  \end{figure}

To produce a graph-structured output with Edge Transformer, a linear
layer with cross-entropy loss can be used for every edge $(i, j)$. For
classification problems with a single label for the entire input the label can be
associated with one of the edges, or edge representations can be
pooled. 

Lastly, an encoder-decoder version of Edge Transformer can be used to perform sequence-to-sequence transduction. Our approach follows that used
with the original Transformer architecture
\citep{vaswani_attention_2017}. First, an encoder Edge Transformer
processes the question and produces a 3D tensor state $x_{enc}$. Next,
the decoder Edge Transformer generates output tokens one by one,
left-to-right. To produce the output distribution for the token we
feed the state of the loop edge $(i,i)$ through a linear layer. To
connect the decoder to the encoder we insert $x_{enc}$ into the
decoder’s initial state. This operation is the Edge Transformer’s
equivalent to how the Transformer decoder cross-attends to the
Transformer encoder’s hidden states. To compute training
log-likelihoods we adapt the Transformer decoder’s causal masking
approach to our triangular attention mechanism.

\section{Experiments}
\label{sec:experiments}

In our experiments we compare the systematic generalization
ability of Edge Transformers to that of Transformers
\citep{vaswani_attention_2017}, Universal Transformers \citep{dehghani_universal_2019}, Relation-aware Transformers
\citep{shaw_self-attention_2018}, Graph Attention Networks
\citep{velickovic_graph_2018} and other baselines. We focus on three synthetic benchmarks
with carefully controlled train-test splits, Compositional Language
Understanding and Text-based Relational Reasoning (CLUTRR), proposed
by \citet{sinha_clutrr_2019}, Compositional Freebase Questions
(CFQ) proposed by \citet{keysers_measuring_2020} and Compositional Generation Challenge based on Semantic Interpretation (COGS) by \citet{kim_cogs_2020}. Systematic
generalization refers to the ability to recombine known primitive
units in novel combinations. These three benchmarks match test and train in
terms of primitive atoms used and rules of combination, but test on
novel combinations of atoms not seen in train. These benchmarks are
thus appropriate for assessing whether Edge Transformers have a
greater ability to generalize systematically than existing models.

In all tables we report mean $\pm$ standard error over multiple runs. For details on the hyperparameter search procedure see Appendix A. For the chosen hyperparameter settings see Table \ref{tbl:hyperparams}.

\begin{table}[]
    \small
    \centering
    \caption{Hyperparameter settings for the Edge Transformer and for
      the baselines. $L$ is the number of layers, $d$ is the dimensionality, $h$ is the number of heads, $B$ is the batch size, $\rho$
      is the learning rate, $T$ is training duration. For CFQ, ``dep.'' stands for dependency parsing and ``sem.'' stands for semantic parsing. }
    \begin{tabular}{lccccccc}
    \toprule
         Model & Task & $L$ & $d$ & $h$ & $B$ & $\rho$ & $T$ \\
    \midrule
         Edge Transformer & CLUTRR & 8 & 200 & 4 & 400 & $1\cdot10^{-3}$ & 50 epochs \\
         RAT & CLUTRR & 8 & 320 & 8 & 200 & $1\cdot10^{-3}$ & 50 epochs \\
         RRAT & CLUTRR & 6 & 320 & 8 & 200 & $1\cdot10^{-3}$ & 50 epochs \\
         Edge Transformer & CFQ dep. & 7 & 360 & 4 & $5 \cdot 10^{2}$ words & $1\cdot10^{-3}$ & 8000 steps \\
         Transformer & CFQ dep. & 7 & 360 & 4 & $1 \cdot 10^{3}$ words & $1\cdot10^{-3}$ & 8000 steps \\
         Universal Transformer & CFQ dep. & 8 & 360 & 4 & $5 \cdot 10^{2}$ words & $1\cdot10^{-3}$ & 8000 steps \\
         Edge Transformer & CFQ sem. & 6 & 256 & 8 & 64 & $6 \cdot 10^{-4}$ & 100 epochs\\
         Universal Transformer & CFQ sem. & 4 & 256 & 8 & 64 &  $6\cdot10^{-4}$ & 100 epochs \\
         Edge Transformer & COGS & 3 & 64 & 4 & 100 &  $5\cdot10^{-4}$ & 200 epochs \\
    \bottomrule
    \end{tabular}
    \label{tbl:hyperparams}
\end{table}

\subsection{Relational Reasoning on Graphs} 
In the first round of experiments, we use the CLUTRR benchmark
proposed by \cite{sinha_clutrr_2019}. CLUTRR  evaluates the 
ability of
models to infer unknown familial relations between individuals 
based on sets of given relations. For example, if one knows that A
is a son of B, C is a son of B and C has an uncle called D, one can
infer that D must also be the uncle of A (see Figure~\ref{fig:tasks},
A). The authors propose training models on scenarios that
require a small number (2, 3 or 4) of supporting relations to infer
the target relation and test on scenarios where more (up to 10)
supporting relations are required. We will refer to the number of
inferential steps required to prove the target relation as the
\textit{relation length} $k$. For relation lengths $k=2,3,4$, we train models on the original CLUTRR
training set from \citet{sinha_clutrr_2019}, which contains 15k examples. For relation lengths $k=2,3$, we generate a larger training set containing $35k$ examples using \citet{sinha_clutrr_2019}'s original code, allowing us to measure systematic generalization performance with less variance. 
We use the noiseless (only the required facts are included)
graph-based version of the challenge, where the input is represented
as a labeled graph, and the model's task is to label a missing edge in
this graph. 

\begin{figure}
\centering
\begin{minipage}{.5\textwidth}
  \centering
  \includegraphics[width=1.1\textwidth]{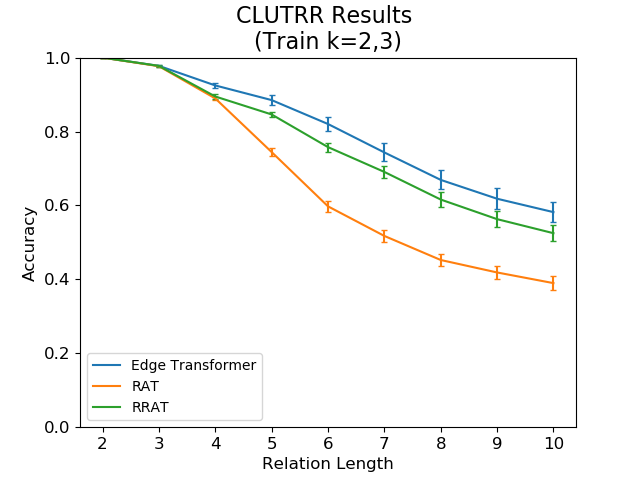}
\end{minipage}%
\begin{minipage}{.5\textwidth}
  \centering
  \includegraphics[width=1.1\textwidth]{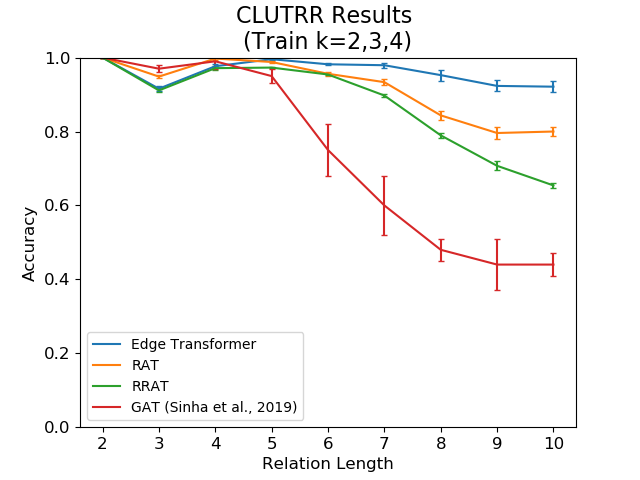}
\end{minipage}
\caption{Relation prediction accuracy on CLUTRR for test relations of
  lengths $k \in [2, 10]$. GAT results from
  \cite{sinha_clutrr_2019} (only shown for the $k=2,3,4$ task, as the $k=2,3$ training set was regenerated). Edge, Relation-Aware (RAT), Relation-updating Relation-aware (RRAT) Transformer 
  results are averaged over 10 runs.}
\label{fig:clutrr}
\end{figure}

We create the Edge Transformer's initial state $X^0$ by embedding edge
labels. We apply a linear layer to the final representation $x^L_{ij}$
of the query edge $(i, j)$ in order to produce logits for a
cross-entropy loss. Our main baseline is a Relation-aware Transformer (RAT)
\citep[RT;][]{shaw_self-attention_2018}, a Transformer variant that
can be conditioned on arbitrary graphs. Edges for RAT are initialized
by embedding edge labels, and nodes are initialized with zero
vectors. Logits are computed by applying a linear layer to
concatenated representations of the nodes that occur in the queried
edge.  We also compare against a variant of RAT which updates both nodes and edges. For this
Relation-updating Relation-aware Transformer (RRAT), node updates are
computed in the same way as in RAT. The update for the edge $x_{ij}$
is computed by concatenating the representations for nodes $i$ and
$j$, and applying a linear layer. Separate feed-forward layers are
applied to nodes and edges. Logits are computed by applying a linear layer to the queried edge.
Finally, for the $k=2,3,4$ setting we include the Graph Attention Networks (GAT,
\citet{velickovic_graph_2018}) results from the original CLUTRR paper
for comparison.
Figure~\ref{fig:clutrr} displays the results. One can see the Edge
Transformer beats all baselines by wide margins, in both $k=2,3$
and $k=2,3,4$ training settings.

\subsection{Dependency Parsing of CFQ}\label{sec:dependency}
In our second set of experiments we use the dependency parsing version
of the CFQ challenge \citep{keysers_measuring_2020} recently proposed
by \citet{goodwin2021compositional}. The original CFQ challenge
evaluates compositional generalization in semantic parsing---in the
case of CFQ, the task of translating natural language questions into
corresponding knowledge graph queries.  The CFQ benchmark was
constructed by generating test/train splits which minimize the
difference between the test and train distributions over primitive
units, like words, while maximizing the \textit{compound
  divergence}---the dissimilarity between test and train distributions
over larger structures, like phrases. As a result, at test time, a
model has to generalize to novel syntactic constructions, such as
``Was NP NP?'' (``Was Alice the sister of Bob?'') that were not seen
during training. \citet{keysers_measuring_2020} released three
different maximum compound divergence (MCD) splits.
\citet{goodwin2021compositional} propose a dependency parsing version
of the CFQ task and show that the MCD splits are challenging for a
state-of-the-art dependency parser (see Figure~\ref{fig:tasks}, B). This dependency parsing task is convenient for evaluating Edge Transformer in an encoder-only graph prediction setting that is simpler than the original sequence-to-sequence formulation of the CFQ task (see Section \ref{sec:cfq_sem} for results in the original setting).

We use the Stanza framework for dependency parsing
\citep{qi_stanza_2020} in which we replace the default BiLSTM model
with Transformer, Universal Transformer or Edge Transformer variants. The models are
trained to label each pair of positions $i$ and $j$ with a dependency
type label or a null label. These predictions are then combined into a
parse tree by Stanza's spanning-tree-based parsing algorithm. The
model's initial state $X^0$ is initialized from token and relational
embeddings as described by Equation~\ref{eq:sequence_init}. For Edge
Transformer, pairwise output labels between positions $i$ and $j$ are
predicted directly from edge $x_{ij}$. For the baselines, these labels
are obtained by passing each pair of nodes through a biaffine layer,
following \citet{dozat_deep_2017}.
The training set contains approximately 100K questions.

We report the Labeled Attachment Score (LAS) that assesses partial
correctness (F1) of the predicted dependency trees as well as the
exact match accuracy (EM) that only takes into account completely
predicted trees. As can be seen in Table \ref{tbl:cfq_results}, Edge
Transformer enjoys a sizable performance advantage (17-30\% EM
accuracy) over all baseline models on the MCD2 and MCD3 splits. On
MCD1 split both the BiLSTM and Edge Transformer achieve near-perfect performance, while the classical Transformer lags behind. 

\begin{table}[] 
\small
\centering 
\caption{Dependency parsing accuracy on
    Compositional Freebase Queries (CFQ). EM is Exact Match accuracy,
    LAS is Labeled Attachment Score. BiLSTM results are from
    \cite{cfq_dependencies}. Transformer, Universal Transformer and Edge Transformer results
    are averaged over 10 runs.}
\label{tbl:cfq_results}
\begin{tabular}{lllllll}
\toprule
Model            & \multicolumn{2}{c}{MCD1} & \multicolumn{2}{c}{MCD2} & \multicolumn{2}{c}{MCD3} \\ 
                 & EM       & LAS        & EM       & LAS        & EM      & LAS
                 \\
\midrule
BiLSTM           & 96.6 $\pm$ 1.3        & \textbf{99.6} $\pm$ 0.2     & 71.4 $\pm$ 2.6      & 93.3 $\pm$ 0.9      & 56.8 $\pm$ 2.8      & 90.9  $\pm$ 1.0    \\
Transformer      & 75.3 $\pm$ 1.7        & 97.0 $\pm$ 0.1     & 59.3 $\pm$ 2.7       & 91.8 $\pm$ 0.4     & 48.0 $\pm$ 1.6      & 89.4 $\pm$ 0.3     \\
Universal Transformer      & 80.1 $\pm$ 1.7        & 97.8 $\pm$ 0.2     & 68.6 $\pm$ 2.3       & 92.5 $\pm$ 0.4     & 59.4 $\pm$ 2.0      & 90.5 $\pm$ 0.5     \\
Edge Transformer & \textbf{97.4} $\pm$ 0.8       & 99.4 $\pm$ 0.2     & \textbf{89.5} $\pm$ 0.5       & \textbf{96.8} $\pm$ 0.1  &   \textbf{89.1} $\pm$ 0.2      & \textbf{96.3}  $\pm$ 0.1   \\
\bottomrule
\end{tabular}
\end{table}

\subsection{Semantic Parsing of CFQ}\label{sec:semantic-parsing}
\label{sec:cfq_sem}

We furthermore experiment with the original semantic parsing task from
CFQ. In this task, natural language questions must be translated to
knowledge graph queries in SPARQL query language (see Figure \ref{fig:tasks}, C). We use the encoder-decoder version of Edge Transformer to produce the sequence of query tokens from a sequence of questions tokens.

The $O(n^3)$ computational complexity and memory footprint of Edge Transformer pose a challenge in this round of experiments as the decoder's input  size $n$ is larger than for the other tasks (more than 100 tokens for some examples). To remedy this, we filter out training examples in which the SPARQL queries are more than 50 tokens long. Even after this filtering training an Edge Transformer model on CFQ semantic parsing requires 1-2 days of using 4 NVIDIA V100 GPUs. This makes a larger scale hyperparameter search prohibitively expensive for this model. To tune Edge Transformer, we first perform the hyperparameter search for the computationally cheaper Universal Transformer baseline which we found to outperform the vanilla Transformer in preliminary experiments. For Edge Transformer we only try a few variations on the chosen Universal Transformer configuration (see Appendix A.2 for more details). As suggested in the original CFQ paper \citep{keysers_measuring_2020}, we tune hyperparameters on the random split and keep the compositional generalization splits for testing. We reduce the size of the random split training data from $\sim$ 90K examples to $\sim$ 20K
to increase the sensitivity to different hyperparameter choices. For each MCD split we report mean exact match accuracy for 30 Universal Transformer and 5 Edge Transformer runs. 

The results are shown in Table \ref{tbl:cfq_sem}. Edge Transformer reaches the state-of-the-art performance among the non-pretrained encoder-decoder models. It reaches 24.67 $\pm$ 1.27 \% average MCD accuracy that approximately 3 percentage points ahead of comparable competition. Notably, Edge Transformer also beats Universal Transformer on our filtered  random split (99.08 $\pm$ 0.1\% vs 98.45 $\pm$ 0.08 \% accuracy).

\begin{table}[t] 
\small
\centering
\caption{Exact match accuracy of the produced SPARQL queries for CFQ semantic parsing experiments. $\spadesuit$ and $\diamondsuit$ denote results from \citep{keysers_measuring_2020} and \citep{furrer_compositional_2020} respectively. We convert the 95\% confidence intervals reported in both papers to standard errors. \label{tbl:cfq_sem}}
\begin{tabular}{lccccc}
\toprule
Model           & MCD avg. & MCD1 & MCD2 & MCD3 \\
\midrule
Universal Transformer          
& 21.26 $\pm$ 1.06 
& 42.7 $\pm$ 2.55 
& 9.46 $\pm$ 1.2
& 11.62 $\pm$ 0.68   \\  
\textbf{Edge Transformer}
& \textbf{24.69} $\pm$ 1.27 
& \textbf{47.73 $\pm$ 3.63}
& \textbf{13.14 $\pm$ 2.12}
& \textbf{13.2 $\pm$ 1.38}  \\    
Universal Transformer $\spadesuit$
& 18.9 $\pm$ 1.1  
& 37.4 $\pm$ 1.8
& 8.1 $\pm$ 1.3
& 11.3 $\pm$ 0.2   \\
Evolved Transformer $\diamondsuit$
& 20.8 $\pm$ 0.6 
& 42.4 $\pm$ 0.8 
& 9.3 $\pm$ 0.6
& 10.8 $\pm$ 0.2   \\
T5-small $\diamondsuit$
& 21.4 $\pm$ 1.2 
& 42.5 $\pm$ 2.1
& 11.2 $\pm$ 1.2
& 10.6 $\pm$ 0.3   \\ 
\bottomrule
\end{tabular}
\end{table}

\subsection{Semantic Parsing of COGS}

As our final evaluation of the ability of Edge Transformers to
generalize systematically, we make use of the COGS benchmark of
\citet{kim_cogs_2020}. COGS defines a semantic parsing task from
natural language sentences to linguistic denotations expressed in a
logical formalism derived from that of \citet[][]{reddy.s:2017}. The
training/test split in COGS is designed to test the ability of models
to generalize from observed sentences, to sentences of greater
complexity that exhibit novel combinations of primitive words and
phrases.  We train and test on the graph
representation of the COGS semantic parses provided by
\citet{ontanon2021making}.
The nodes in these semantic graphs are input tokens that are labeled with word-level category information, for
example, semantic role and part-of-speech. The edges connect some pairs of nodes to indicate
parent-child dependencies between them, similar to a dependency
parse (see Figure \ref{fig:tasks},D). The Edge Transformer model predicts node labels for token $i$
by applying several linear layers (one for each label type) to
self-edges $x_{ii}$. For edge labels, each node has a unique parent
(for nodes with no parent, we follow \citet{ontanon2021making} and
treat them as being their own parent). To predict the parent of node
$i$, we apply a linear layer to all edges $x_{ji}$, and take a softmax
across the position axis.  Hyperparameters for Edge Transformer were
not tuned for this task. Instead, we used the best setting identified by \citet{ontanon2021making} for the Transformer
architecture (see Appendix A.3), and default settings for the optimizer.


As can be seen from Table~\ref{table:cogs}, Edge Transformers exhibits
a state-of-the-art performance of 87.4 $\pm$ 0.4~\% on COGS,
outperforming the baseline graph prediction models from
\citep{ontanon2021making} by 9 percentage
points on this challenging task. This performance is also higher than
what is reported in the literature for the original sequence
prediction version of the COGS task, including 82\% accuracy of the
lexicon-equipped LSTM (LSTM+Lex by \citet{akyurek_learning_2021}) and
81\% accuracy of the carefully initialized Transformer model
\citep{csordas_devil_2021}.

\begin{table}[] 
\small
\centering
\caption{Top section: graph prediction accuracy for Edge Transformer
  and comparable graph prediction Transformer results by
  \citet{ontanon2021making} (marked by $\diamondsuit$). Bottom
  section: best COGS results by \citet{akyurek_learning_2021}
  ($\spadesuit$) and \citet{csordas_devil_2021} ($\clubsuit$) obtained
  by predicting semantic parses as sequences of tokens.}
\label{table:cogs}
\begin{tabular}{ll}
\toprule
Model            & Generalization Accuracy        
                 \\
\midrule
\textbf{Edge Transformer}      & \textbf{87.4} $\pm$ 0.4    \\    
Universal Transformer - Attention Decoder $\diamondsuit$ & 78.4          \\
\hline\hline
LSTM+Lex $\spadesuit$ & 82 $\pm$ 1. \\
Transformer $\clubsuit$ & 81 $\pm$ 0.01 \\
\bottomrule
\end{tabular}
\end{table}

\subsection{Ablation Experiments}

To explore the performance of Edge Transformers, we experiment with
several variations of the model. First, we train Edge Transformers
with untied weights, using separate weights for each layer, as is
usually done with classical Transformers. We refer to this variant as
\textit{untied weights}.  Second, we simplify the Edge Transformer
computations by computing the value $v_{ilj}$ for each triangle
$(i, l, j)$ using only the edge $(i, l)$ instead of using both edges
$(i, l)$ and $(l, j)$.  This is equivalent to replacing Equation
\ref{eq:edge_attention_end} with $v_{ilj} = V^1 x_{il}$. We will refer
to this condition as \textit{value ablation}. Third and finally, we use
a key vector $k_{ij} = W^k x_{ij}$ instead of $k_{lj} = W^k x_{lj}$
for computing attention scores in Equation
\ref{eq:edge_attention_begin}. We call this condition \textit{attention ablation}.

We compare our base model to the above variants on the CLUTRR task, 
the MCD3 split of the CFQ dependency parsing task and on the COGS task. In CLUTTR experiments we rerun the hyperparameters for each
ablated model, whereas in other experiments we keep the same
hyperparameters as for base Edge Transformer. Table \ref{tbl:lesion_clutrr} shows results for CLUTRR and Table 
\ref{tbl:lesion_cfq} shows results for CFQ and COGS. One can see that value ablation consistently
leads to worse performance on CLUTRR and CFQ dependency parsing, confirming the importance of the
unification-inspired triangular updates described in Equations
\ref{eq:edge_attention_begin}-\ref{eq:edge_attention_end} and
illustrated in Figure \ref{fig:edge_transformer}. Likewise, attention ablation hurts the performance on CLUTRR and COGS.
Lastly, for untying weights,
there is some evidence of deterioration for the CLUTRR setting where
we train on $k=2,3$, and a large effect for CFQ dependency parsing.

\begin{table}
\small
\caption{
  Results of ablation experiments on CLUTRR. ``Base'' is
  Edge Transformer as described in Section \ref{sec:edge_transformer}.
  See Section \ref{sec:experiments} for details on ablated model
  variants. Results are averaged over 10 runs. }
    \label{tbl:lesion_clutrr}
    \centering
    \begin{tabular}{cllllll}
    \toprule
    \multicolumn{1}{l}{}                                                       & Model                    & \multicolumn{5}{c}{Test k} \\ 
    
    \multicolumn{1}{l}{}                                                       &                          & 6       & 7      & 8      & 9      & 10     \\ 
    \midrule
    Train on k=2,3
    & Base   & 82.0 $\pm$ 2.0    & 74.3 $\pm$ 2.4   & 66.9 $\pm$ 2.6 & 61.8 $\pm$ 2.9   & 58.1 $\pm$ 2.8  \\
 & Untied weights        & 78.9 $\pm$ 1.0    & 71.0 $\pm$ 1.1   & 63.2 $\pm$ 1.1   & 58.5 $\pm$ 1.2   & 55.3 $\pm$ 1.2   \\
  & Value ablation & 54.4 $\pm$ 3.4    & 44.2 $\pm$ 3.6   & 36.9 $\pm$ 3.6   & 33.9 $\pm$ 3.7   & 30.7 $\pm$ 3.6   \\ 
  & Attention ablation & 84.4 $\pm$ 1.7   & 73.1 $\pm$ 2.3   & 63.2 $\pm$ 2.2   & 56.4 $\pm$ 2.1  & 51.4 $\pm$ 1.9  \\
                                                                                \hline
    Train on k=2,3,4 
    & Base         & 98.1 $\pm$ 0.3   & 97.9 $\pm$ 0.7  & 95.3  $\pm$ 1.5   & 92.3 $\pm$ 1.5   & 92.1 $\pm$ 1.5   \\
    & Untied weights    & 98.2 $\pm$ 0.3   & 98.2 $\pm$ 0.3   & 95.6 $\pm$ 0.7   & 93.1 $\pm$ 1.4  & 90.2 $\pm$ 1.4  \\
    & Value ablation & 92.6 $\pm$ 1.3    & 86.2 $\pm$ 2.1   & 82.5 $\pm$ 3.2  & 77.1 $\pm$ 3.7  & 75.2 $\pm$ 2.7  \\
     & Attention ablation & 95.9 $\pm$ 2.1   & 95.2 $\pm$ 2.5  & 92.8 $\pm$ 3.1   & 89.8 $\pm$ 2.9  & 78.7 $\pm$ 2.5  \\
    \bottomrule
    \end{tabular}
\end{table}
\begin{table}
\small
\caption{Results of ablation experiments on CFQ dependency parsing and COGS graph semantic parsing. ``Base'' stands for the Edge Transformer as described in
  Section \ref{sec:edge_transformer}. See Section
  \ref{sec:experiments} for details on ablated model variants.
  Results are averaged over 10 runs. EM is Exact Match accuracy, LAS
  is Labeled Attachment Score. }
    \label{tbl:lesion_cfq}
    \centering
    \begin{tabular}{lccc}
    \toprule
    Model                    & \multicolumn{2}{c}{MCD3} & COGS\\
    
                             & EM      & LAS & Gen. Accuracy        \\
    \midrule
    Base         & 89.1 $\pm$ 0.2      & 96.3  $\pm$ 0.1 & 87.4 $\pm$ 0.4 \\
    Untied weights           & 37.2 $\pm 12.4$ & 53.2 $\pm 14.2$ & 86.3 $\pm$ 0.8     \\
    Value ablation & 84.5 $\pm 1.1 $      & 95.5 $\pm 0.3$ & 87.8 $\pm$ 0.5 \\
    Attention ablation & 88.1 $\pm 0.7$ & 96.4 $\pm 0.1$ & 86.3 $\pm$ 0.7 \\
    \bottomrule
    \end{tabular}
\end{table}

\section{Related Work}
Most variations of the Transformer architecture by
\citet{vaswani_attention_2017} aim to improve upon its quadratic
memory and computation requirements. To this end, researchers have
experimented with recurrency \citep{dai_transformer-xl_2019}, sparse
connectivity \citep{ainslie_etc_2020}, and kernel-based approximations
to attention \citep{choromanski_rethinking_2021}. Most relevant here
is the use of relative positional embeddings in Relation-aware
Transformers proposed by \citep{shaw_self-attention_2018} and refined
by \citep{dai_transformer-xl_2019}. While this approach introduces
edges to Transformers, unlike in Edge Transformers, these edges are
static; that is, they do not depend on external input other than the
respective edge labels.

Graph Neural Networks
\citep[GNNs;][]{gori_new_2005,scarselli_graph_2009,
  kipf_semi-supervised_2017}, and especially Graph Attention Networks
\citep{velickovic_graph_2018} are closely related to Transformers,
especially relation-aware variants. Unlike Transformers, GNNs operate
on sparse graphs. While some GNN variants use edge features
\citep{gilmer_neural_2017,chen_utilizing_2019} or even compute dynamic
edge states \citep{gong_exploiting_2019}, to the best of our knowledge
no GNN architecture updates edge states based on the states of other
edges as we do in
Equations~\ref{eq:edge_attention_begin}-\ref{eq:edge_attention_end}.

A number of studies of systematic generalization have examined the
ability of neural networks to handle inputs consisting of familiar
atoms that are combined in novel ways. Such research is
typically conducted using carefully controlled train-test splits
\citep{lake_generalization_2018,hupkes_compositionality_2019,sinha_clutrr_2019,bahdanau_systematic_2019,keysers_measuring_2020,kim_cogs_2020}.
Various techniques have been used to improve generalization on some of
the these benchmarks, including (but not limited to) data augmentation
\citep{andreas_good-enough_2019,akyurek_learning_2020}, program
synthesis \citep{nye_learning_2020}, meta-learning
\citep{lake_compositional_2019}, and task-specific model design
\citep{guo_hierarchical_2020}. Most relevant to our work is the Neural
Theorem Proving (NTP) approach proposed by
\citep{rocktaschel_end--end_2017} and developed by
\citep{minervini_learning_2020}. NTP can be described as a
differentiable version of Prolog backward-chaining inference with
vector embeddings for literals and predicate names. Unlike NTP, Edge
Transformers do not make rules explicit; we merely provide the
networks with an architecture that facilitates the learning of
rule-like computations by the attention mechanism. Edge Transformers
remain closer to general purpose neural models such as Transformers in
spirit and, as demonstrated by our CFQ and COGS experiments, can be
successfully applied in the contexts where the input is not originally
formulated as a knowledge graph.

Finally, the Inside-Outside autoencoder proposed by
\citet{drozdov_unsupervised_2019} for unsupervised constituency
parsing is the most similar to our work at the technical level. The
Inside-Outside approach also features vector states for every pair of
input nodes. However, in contrast to the Edge Transformer, the
attention is restricted by the span structure of the input.

\section{Discussion}
\label{sec:discussion}
We present the Edge Transformer, an instantiation of the idea that a
neural model may benefit from an extended 3D-tensor state that more
naturally accommodates relational reasoning. Our experimental results
are encouraging, with clear performance margins in favor of the Edge
Transformer over competitive baselines on three compositional
generalization benchmarks. Our ablations experiments confirm the importance of core
intuitions underlying the model design.

Edge Transformers can be seen as belonging to the line of work on
neuro-symbolic architectures \citep[][to name but a
few]{mao_neuro-symbolic_2018,rocktaschel_end--end_2017,evans_learning_2018}.
Compared to most work in the field, Edge Transformers are decidedly
more neural: no logical gates or rule templates can be found in our
architecture. We believe this is a promising direction for further
exploration, as it allows the model to retain the key advantages of
neural architectures: their generality, interchangeability and ability
to learn from data.

Edge Transformer computations and memory usage scale as $O(n^3)$ with
respect to the input size~$n$. While this makes scaling to larger
inputs challenging, in our experience modern GPU hardware allows
training Edge Transformers with inputs of size up to a hundred
entities. One way to use Edge Transformer in applications with larger
inputs is to create Edge Transformer nodes only for the most salient
or important components of an input. For example, in the language
modelling context these could be just the heads of noun phrases and/or
just the named entities. Future work could build and pretrain on raw
text hybrid models that combine both Transformer- and Edge
Transformer-based processing, the latter performed only for a small
subset of nodes.

\newpage

\begin{ack}
  We gratefully acknowledge the support of NVIDIA Corporation with the donation of two Titan V GPUs used for this research. 


\end{ack}


\bibliography{references,extra_references}
\bibliographystyle{apalike}

\newpage

\appendix
\section{Additional Experiment Details}
\label{sec:hpsearch}

\subsection{CLUTTR and CFQ Dependency Parsing}
\label{sec:cluttr_and_cfqdep_hpsearch}
The experiments were performed using a cluster of 12 GPUs (2 24GB, 2
12GB, 8 11GB). A single training run on CFQ or CLUTRR requires less
than 30 minutes on a single GPU. Hyperparameters were selected using
grid search. For both the Edge Transformer and (Relation-aware)
Transformer models, hyperparameter search was the same. The number of units
were varied between 200-400; batch size between 40 and 400 (for
CLUTRR) and $5\cdot 10^3$ and $1 \cdot 10^4$ words (for CFQ).
Learning rates varied between $1\cdot10^{-4}$ and $2 \cdot
10^{-3}$, and the number of heads varied from
between 4 and 8. For Transformer models, the number of layers varied from 5 to 8. For
Edge Transformer models, the number of layers was varied from 6 to 8.

Hyperparameters for CLUTRR (for ET, baselines, and ablations) were optimized on the $k=2,3$ task, and fixed for the $k=2,3,4$ task. For CFQ dependency parsing, hyperparameters were optimized on a $1\%$ random subset of the official random split (Keysers et al., 2020), and fixed for the MCD splits. 

\subsection{CFQ Semantic Parsing}
\label{sec:cfqsem_hpsearch}
When tuning the Universal Transformer baseline we varied the number of layers from 2 to 6, the learning rate from 0.0001 to 0.001, the batch size of 64 to 128 and the number of epochs on the 20K random split training examples from 200 to 400. The number of heads was fixed to 8. For training on MCD splits with $\sim$90K training examples each we divided the number of epochs by 4 to keep the total number of training steps approximately the same. We found that longer training gave better results, hence all results in the paper are obtained with 100 training epochs.

\subsection{COGS Semantic Parsing}
\label{sec:cogs_hpsearch}
No hyperparameter search was performed for Edge Transformer on COGS. Architecture hyperparameters for Edge Transformer were matched to those of (Ontanón et al., 2021), who tuned the number of
  layers, hidden dimension, feed-forward dimension, and number of
  heads for their Transformer architectures. Their best Transformer model has two standard layers and a separate attention head for output, giving three QK attention layers total. We therefore use three layers for Edge Transformer. Default settings were used for optimizer hyperparameters.



\section*{Checklist}


\begin{enumerate}

\item For all authors...
\begin{enumerate}
  \item Do the main claims made in the abstract and introduction accurately reflect the paper's contributions and scope?
    \answerYes{}
  \item Did you describe the limitations of your work?
    \answerYes{See Section 6.}
  \item Did you discuss any potential negative societal impacts of your work?
    \answerNo{}
  \item Have you read the ethics review guidelines and ensured that your paper conforms to them?
    \answerYes{}
\end{enumerate}

\item If you are including theoretical results...
\begin{enumerate}
  \item Did you state the full set of assumptions of all theoretical results?
    \answerNA{}
	\item Did you include complete proofs of all theoretical results?
    \answerNA{}
\end{enumerate}

\item If you ran experiments...
\begin{enumerate}
  \item Did you include the code, data, and instructions needed to reproduce the main experimental results (either in the supplemental material or as a URL)?
    \answerYes{}
  \item Did you specify all the training details (e.g., data splits, hyperparameters, how they were chosen)?
    \answerYes{}
	\item Did you report error bars (e.g., with respect to the random seed after running experiments multiple times)?
    \answerYes{}
	\item Did you include the total amount of compute and the type of resources used (e.g., type of GPUs, internal cluster, or cloud provider)?
    \answerYes{}
\end{enumerate}

\item If you are using existing assets (e.g., code, data, models) or curating/releasing new assets...
\begin{enumerate}
  \item If your work uses existing assets, did you cite the creators?
    \answerYes{}
  \item Did you mention the license of the assets?
    \answerNo{}
  \item Did you include any new assets either in the supplemental material or as a URL?
    \answerNo{}
  \item Did you discuss whether and how consent was obtained from people whose data you're using/curating?
    \answerNA{}
  \item Did you discuss whether the data you are using/curating contains personally identifiable information or offensive content?
    \answerNA{}
\end{enumerate}

\item If you used crowdsourcing or conducted research with human subjects...
\begin{enumerate}
  \item Did you include the full text of instructions given to participants and screenshots, if applicable?
    \answerNA{}
  \item Did you describe any potential participant risks, with links to Institutional Review Board (IRB) approvals, if applicable?
    \answerNA{}
  \item Did you include the estimated hourly wage paid to participants and the total amount spent on participant compensation?
    \answerNA{}
\end{enumerate}

\end{enumerate}

\end{document}